\documentclass[runningheads]{llncs}
\usepackage[T1]{fontenc}
\usepackage{graphicx}

\usepackage{amssymb}
\usepackage{color}
\usepackage[dvipsnames,table,xcdraw]{xcolor}
\usepackage{amsmath}
\usepackage{amsfonts}
\usepackage{booktabs}
\usepackage[ruled,linesnumbered,vlined]{algorithm2e}
\usepackage{multirow}
\usepackage{pifont}

\SetKwInput{KwInput}{Input}                %
\SetKwInput{KwOutput}{Output}              %

\def\eg{\textit{e.g., }}
\def\ie{\textit{i.e., }}

\begin{document}
\title{AdaCBM: An Adaptive Concept Bottleneck Model for Explainable and Accurate Diagnosis}

\titlerunning{AdaCBM}

\author{Townim F. Chowdhury \inst{1} \and
Vu Minh Hieu Phan \inst{1} \and
Kewen Liao \inst{2} \and
Minh-Son To \inst{3} \and
Yutong Xie \inst{1} \and
Anton van den Hengel \inst{1} \and
Johan W. Verjans \inst{1} \and
Zhibin Liao \inst{1}\thanks{Corresponding author}
}
\authorrunning{TF.Chowdhury et al.}
\institute{Australian Institute for Machine Learning, University of Adelaide, Australia \and
Australian Catholic University, Australia \and
Flinders University, Australia
}

\maketitle              %
\begin{abstract}
The integration of vision-language models such as CLIP and Concept Bottleneck Models (CBMs) offers a promising approach to explaining deep neural network (DNN) decisions using concepts understandable by humans, addressing the black-box concern of DNNs. While CLIP provides both explainability and zero-shot classification capability, its pre-training on generic image and text data may limit its classification accuracy and applicability to medical image diagnostic tasks, creating a transfer learning problem. 
To maintain explainability and address transfer learning needs, CBM methods commonly design post-processing modules after the bottleneck module. However, this way has been ineffective.
This paper takes an unconventional approach by re-examining the CBM framework through the lens of its geometrical representation as a simple linear classification system. The analysis uncovers that post-CBM fine-tuning modules merely rescale and shift the classification outcome of the system, failing to fully leverage the system's learning potential.
We introduce an adaptive module strategically positioned between CLIP and CBM to bridge the gap between source and downstream domains. This simple yet effective approach enhances classification performance while preserving the explainability afforded by the framework. Our work offers a comprehensive solution that encompasses the entire process, from concept discovery to model training, providing a holistic recipe for leveraging the strengths of GPT, CLIP, and CBM. Code is available at: \textbf{\texttt{\url{https://github.com/AIML-MED/AdaCBM}}}.

\keywords{Explainable Diagnosis \and Interpretability  \and Concept Bottleneck Model \and Model Fine-tuning.}
\end{abstract}

\section{Introduction}
Despite the rapid development of medical AI, distrust among healthcare practitioners and the general public hinders the deployment of AI systems in clinical practice. 
The demand for transparent AI reasoning~\cite{lipton2017doctor} has led to the emergence of explainable AI (XAI) research. 
Visualizing saliency maps~\cite{liao2020multi,brunese2020explainable,xie2020mutual} and class activation maps~\cite{zhou2016learning,selvaraju2017grad} are typical XAI methods, although several research~\cite{adebayo2018sanity,rudin2019stop} pointed out that the visualization can be inaccurate and difficult to act on.
Concept Bottleneck Model (CBM) \cite{koh2020concept} interprets deep learning models in language-described terms (namely text concepts, or simply, concepts) and shows how decision weighs on those concepts. 
CBM allows users to judge and exclude non-relevant concepts to adjust the model prediction at inference time, which can help clinicians understand and build trust in such an AI system. 
Nevertheless, applying CBM is challenging in medical image diagnosis due to the time-consuming and costly process of labeling medical concepts on images. 

Recently, several `label-free' CBM approaches~\cite{oikarinen2023label,yang2023language,panousis2023sparse} utilize Large Language Models (LLMs; \eg GPT \cite{brown2020language}) to generate disease attributes and subsequently use them as concepts to obtain concept-image association scores from Vision-Language Models (VLMs; \eg CLIP \cite{radford2021learning}).
Despite commendable interpretability capability, the label-free CBM methods face three main challenges in medical image classification tasks. \textbf{(1)} CLIP was pre-trained on images and texts in the general domain, which has low zero-shot capability in medical domain tasks. %
\textbf{(2)} CBM generally needs a large concept base to perform well, but that would reduce the concept weight sparsity and decrease the model interpretability as shown in~\cite{espinosa2022concept,havasi2022addressing}. %
\textbf{(3)} %
Generating and selecting accurate medical concepts for pathological classification is practically challenging.
Previous works~\cite{yang2023language,oikarinen2023label} employ rule-based concept filtering/trimming~\cite{oikarinen2023label} and concept shortening (\eg using a secondary language model~\cite{yang2023language}) to post-process GPT generated concepts, which are difficult to generalize for non-technical users. 

Addressing the first two challenges together, this work first re-frames the CBM framework to a linear classification system.
We identify that the ``inputs to the system'' should be trained in order to improve the overall fine-tuning ability while adequately preserving CLIP's representation power and CBM's interpretability.
Then, we design a simple yet effective learnable adapter module placed between CLIP and CBM, enabling high classification performance even with a low number of concepts.  
For the last challenge, we design a fully prompt engineering-based concept generation to control concept conciseness and generated disease characteristics (visual appearances, \eg color and shape). 
We also design a \textit{concept utility selection} method that employs established statistical tests to evaluate concepts against the downstream task utility.

In summary, our contributions are as follows.
(1) We re-examine CBM as a linear classification system and derive the location between CLIP and CBM models as the optimal location to add model capacity to mitigate the CLIP and downstream task domain gap. 
(2) We design a fully prompt engineering-based concept generation strategy utilizing GPT-4, which allows users to generate clinically relevant concepts adhering to targeted concept conciseness and visual attributes of the target diseases (\eg color and shape). 
(3) Based on the $t$-statistics and Pearson's $r$ tests, we develop sensible concept selection criteria that maximize concept utility toward the downstream medical diagnostic task.

\section{Methodology}

\noindent \textbf{CLIP-based concept bottleneck model.} CLIP offers a well-pretrained basis to construct label-free concept bottleneck models (CBM)~\cite{koh2020concept}.
Let $\mathbf{x}, \mathbf{t} \in \mathbb{R}^d$ be the image and text embeddings after being processed by respective encoders, CLIP matches the image and language embeddings by their cosine similarity:
    $s = \text{cos}(\mathbf{x}, \mathbf{t}) = \frac{\mathbf{x} \cdot \mathbf{t}}{\lVert\mathbf{x}\rVert \lVert\mathbf{t}\rVert}$.
By defining $\mathbf{S} = [s_1, \ldots, s_K]^\intercal \in \mathbb{R}^{K}$ which contains the cosine similarity scores for $K$ concepts $\mathcal{T} = \{1, \ldots, \mathbf{t}_K\}$ and $\mathcal{C} = \{c_1, \ldots, c_n\}$ a set of $n$ classes, we show the class inference mechanism using CBM as $p(\mathcal{C}|\mathbf{x}, \mathcal{T
}) = \text{softmax}(\mathbf{W}^{\intercal}\mathbf{S} + \mathbf{b})$, where $\mathbf{W} \in \mathbb{R}^{K \times n}$ is a class-concept weight matrix~\cite{yang2023language} and $\mathbf{b} \in \mathbb{R}^{K}$ denotes class bias.
Ignoring the softmax for simplicity, we show a class logit is computed by:
\begin{equation}
    \mathbf{z}_i = \sum_{j=1}^K \mathbf{W}_{ji} s_j + \mathbf{b}_i.
\label{eq:logit_cbm}
\end{equation}
To interpret the CBM, the individual $\mathbf{W}_{ji} s_j$ represents the concept contribution of $\mathbf{t}_j$ toward a class prediction of $c_i$, shown as the red colored bars in Fig.~\ref{fig:da_cbm_example}.

\noindent \textbf{Low adaptation power of CLIP-based CBMs.} In the CLIP-based CBMs, the image and text encoders are typically not fine-tuned~\cite{oikarinen2023label,yang2023language} which leaves the class-concept weight matrix $\mathbf{W}$ and class bias $\mathbf{b}$ the only trainable parameters to fit a down-stream task. 
This will require a large number of $K$ to increase the size of $\mathbf{W}$ to sufficiently learn toward a downstream task but that contradicts the goal of improving human interpretability because interpretability requires fewer activated weight values.
Weight sparsity promotion~\cite{oikarinen2023label,panousis2023sparse} is commonly employed; yet, sparse parameters can restrict the adaptation power of the model
So, the question is where else can we add parameters while maintaining \textbf{(1)} the CBM interpretability %
(\ie CBM is the only inference pathway, instead of creating a residual fitting branch for inference as in~\cite{yuksekgonul2023posthoc})
and \textbf{(2)} maintain the sparsity of the class-concept weight matrix $\mathbf{W}$.
To find an answer to this question, we review CBM from a fresh geometrical representation perspective.
\begin{figure}[!tbp]
    \centering
    \includegraphics[width=1\textwidth]{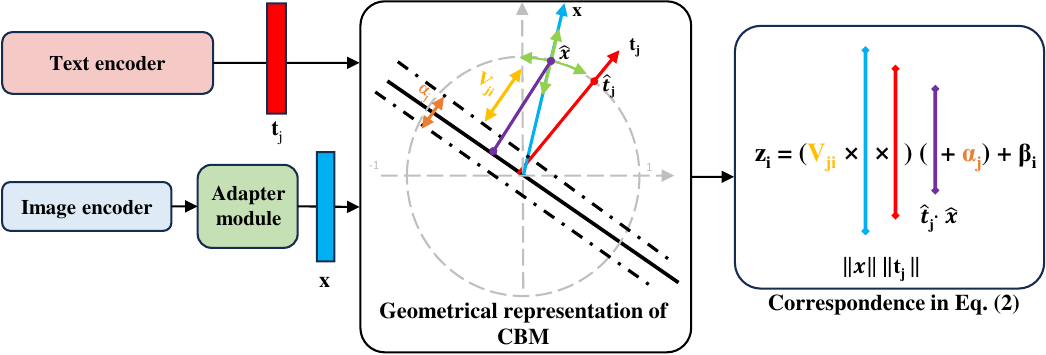}
    \caption{A geometrical representation of the CBM framework. The cosine similarity of an image {\color{Cyan}$\mathbf{x}$} and a text concept {\color{red}$\mathbf{t}_j$} is represented by the distance from the point $\hat{\mathbf{x}}$ on the sphere to the classification boundary (black line), \ie the {\color{Plum} purple $\hat{\mathbf{x}} \cdot \hat{\mathbf{t}}_j$} quantity. Our adapter module provides adaptation ability to {\color{Cyan}$\mathbf{x}$} as indicated by the green arrows.}
\label{fig:geometric_representation}
\end{figure}

\noindent \textbf{CBM is a linear classification system.}
The geometrical representation of a CLIP-based CBM is illustrated in Fig.~\ref{fig:geometric_representation} using a 2D example.
The geometrical meanings of CBM quantities in the system are described below.
\begin{itemize}
    \item $\hat{\mathbf{x}}$ and $\lVert\mathbf{x}\rVert$ are the unit vector and norm of the image embedding $\mathbf{x}$. Since CLIP uses cos similarity, here $\hat{\mathbf{x}}$ is a \textit{point} to be classified by the system, geometrically speaking, it is a $d$-dimensional point that sits on the unit hyper-sphere (gray circle).
    \item $\hat{\mathbf{t}}_j$ and $\lVert\mathbf{t}_j\rVert$ are the unit vector and norm of the text embedding $\mathbf{t}_j$. $\hat{\mathbf{t}}_j$ defines the normal vector of the \textit{classification boundary} (solid black line) derived from a text concept. In simple words, a text encoder here is a projector that projects any text sentence to a classification boundary.
    \item $\hat{\mathbf{x}}\cdot\hat{\mathbf{t}}_j$ is equivalent to the cosine similarity $s_j$ computed by the CLIP model. Geometrically, it defines the point's \textit{classification response}.
    \item $\mathbf{V}_{ji}$ and $\alpha_j$ are post-processing scalar type of \textit{scaling and shift} for any point classified by $\hat{\mathbf{t}}_j$.
    The effects of $\mathbf{V}_{ji}$ and $\alpha_j$ are visualized by the arrows of the respective color in Fig.~\ref{fig:geometric_representation}.
\end{itemize}
Based on the geometrical representation, we reformulate $\mathbf{z}_i$ in Eq.~\ref{eq:logit_cbm} to:
\begin{equation}
    \mathbf{z}_i = \sum_{j=1}^K \mathbf{W}_{ji} s_j + \mathbf{b}_i = \sum_{j=1}^K \big(\mathbf{V}_{ji} \lVert\mathbf{x}\rVert \lVert\mathbf{t}_j\rVert\big) (\hat{\mathbf{x}}\cdot\hat{\mathbf{t}}_j + \alpha_j) + \beta_i,
\label{eq:generalized_cbm}
\end{equation}
where the original class-concept weight $\mathbf{W}_{ji}$ decomposes to $\mathbf{W}_{ji} = \mathbf{V}_{ji} \lVert\mathbf{x}\rVert \lVert\mathbf{t}_j\rVert$. 
For completeness, we introduce an additional bias parameter $\beta \in \mathbb{R}^n$ which can composite the original class bias as $\mathbf{b}_i = \sum_{j=1}^K \big(\mathbf{V}_{ji} \lVert\mathbf{x}\rVert \lVert\mathbf{t}_j\rVert\big) \alpha_j + \beta_i$.
Therefore, we can see that the fine-tuning of CBM class-concept weights only learns the scalars $\mathbf{V}_{ji}$, $\alpha_i$, and $\beta_i$, passively accepting the outcome of the classification system rather than utilizing it. 
Essentially, utilizing the learning capacity of the classification system means changing the input point positions and/or placement of the decision boundary but medical image classification tasks do not often contain paired text data to meaningfully learn the text embedding, so $\mathbf{t}_j$ should ideally be fixed, leaving the input $\mathbf{x}$ the only sensible quantity to change. 
In granular detail, the change can be further decomposed into (1) angular change of $\hat{\mathbf{x}}$ along the surface of the sphere and (2) vector norm $\lVert\mathbf{x}\rVert$, annotated by the green arrows in Fig.~\ref{fig:geometric_representation}.

\noindent \textbf{An adaptive CBM (AdaCBM) to mitigate domain difference.}
The above analysis essentially depicts a transfer learning problem but just tuning the image backbone, which is at risk of overfitting if the full backbone is tuned especially when a large capacity backbone is used.
Therefore, we introduce a learnable module $\mathbf{x} \mapsto F(\mathbf{x}) \in \mathbb{R}^d$ with a controlled number of parameters to mitigate the domain differences.
In essence, AdaCBM brings the image embedding $\mathbf{x}$ closer to the text embedding $\mathbf{t}_j$ by pushing $\mathbf{x}$ toward $\mathbf{t}_j$ if the image contains the concept, or otherwise toward -$\mathbf{t}_j$.
The adapter module is realized by a stacked linear layer with Leaky ReLU~\cite{maas2013rectifier} activation using a 0.01 negative slope setting.
Furthermore, to promote the sparsity in $\mathbf{V}_{ji}$, we follow the execution of LaBo~\cite{yang2023language} to initialize the CBM with a pre-selected $k$ number of concepts for a class $c_i$ (\ie $k=\frac{K}{n}$).
But unlike LaBo which still allows co-adaptation~\cite{srivastava2014dropout} from concept responses for other classes, we assign a stationary mask $\mathbf{M} \in \mathbb{R}^{K\times n}$ which forces the concept contribution only learned from the selected concepts, \ie $\mathbf{M}_{ji} = 1$ if $\mathbf{t}_j$ was selected for $c_i$, otherwise, 0.
To summarize, the logit equation integrated with the AdaCBM is:
\begin{equation}
    \mathbf{z}_i = \sum_{j=1}^K (\mathbf{M} \odot \mathbf{V})_{ji}  \big(F(\mathbf{x})\cdot\mathbf{t}_j + \alpha'_j\big) + \beta_i,
\label{eq:da_cbm}
\end{equation}
where we note $\alpha'_j = \lVert F(\mathbf{x})\rVert \lVert\mathbf{t}_j\rVert \alpha$ for simplicity.
The initial values are set to zeros for $\alpha'_j$ and $\beta_i$.
For $\mathbf{V}$, the initial values are set to $\mathbf{M}$. 
During the training, $F(\mathbf{x})\cdot\mathbf{t}_j$ can be used without explicit normalization to compute the cosine similarity. 
During CBM interpretation, it can be decomposed into the norms and cosine similarity for analysis as shown in Table 2-(1) in the supplementary to illustrate the effectiveness of each term.

\noindent \textbf{Selecting task-relevant medical concepts.}
The feasibility of label-free CBMs lies in the powerful text generation ability of the LLMs (\eg GPTs~\cite{brown2020language}).
With the CLIP's ability to turn any text into embedding, we can virtually generate an indefinite amount of concepts regardless of semantic relevance.
Therefore, concept selection becomes an important step in maintaining the interpretability of a label-free CBM.
To judge the usefulness of a concept, we define the utility of a concept by its discriminability toward a downstream task class, \ie $U(\mathbf{t}, c)$, represented by the \textit{t}-statistic used in Welch's \textit{t}-test:
    $U(\mathbf{t}, c) = \frac{\mu_c - \mu_{c'}}{\sqrt{\frac{\sigma_c^2}{N_c}} + \sqrt{\frac{\sigma_{c'}^2}{N_{c'}}}}$
, where $\mu_c = \sum_{\mathbf{x} \in \mathcal{X}, l=c} \mathbf{x} \cdot \mathbf{t}$ is the mean of dot products where the image embeddings share the same label $l=c$. $\mu_{c'} = \sum_{\mathbf{x} \in \mathcal{X}, l\neq c} \mathbf{x} \cdot \mathbf{t}$ computes the opposite.
Here $\mathcal{X}$ denotes the set of image embeddings in the training set,
$\sigma^2$ and $N$ denote the respective variance and number of samples.
In brief, the \textit{t}-statistic measures whether the population means $\mu_{c}$ and $\mu_{c'}$ are different.
Here we look for concepts that achieve higher responses for the image embeddings of the target class, \ie $\mu_{c} \gg \mu_{c'}$ and detected by  $\text{argmax}_{\mathbf{t}\in\mathcal{T}} U(\mathbf{t}, c)$, from a pool of $\mathcal{T}$ candidate concepts.
When selecting a group of concepts for class $c$, similar concept response patterns will reduce the discriminative power of the group, therefore we introduce Pearson's \textit{r} to reduce the inclusion of highly correlated concepts by a predetermined threshold $\gamma$.
Empirically we found $\gamma = 0.9$ is sufficient to remove the concepts that are worded similarity to each other.
Our concept selection algorithm is illustrated in Algorithm 1 in the supplementary material.

\noindent {\bf Prompt guided medical concept generation.} 
Leveraging the power of GPT-4, we demonstrate the flexibility and effectiveness of prompt engineering in concept generation, we emphasize the concept generation on two aspects: visually represented medical concepts and concept conciseness. For visually represented medical concepts, we prompt GPT-4 to describe a disease class's visual appearance in four categories: color, shape, size, and texture. For example, for generating concepts that represent the color of a disease, we do ``What are the \{colors\} of \{disease name\} can be present in an image of \{disease name\}?''. To constrain concept conciseness, we prompt GPT-4 to make concepts in different word lengths, 3 to 4 words, 5 to 6 words, and 8 to 10 words. We add to the above prompt by commanding, \eg ``each sentence will be \{8\}-\{10\} words.'' On average, we found the average returned concept word lengths are 4.4, 6.0, and 8.7 respectively. This utility helps us to understand the task relevance of the selected concepts. For example in Table 1-Left in the supplementary material, a noticeably higher proportion for shape and texture-based concepts was selected for the HAM dataset while the preference of concept conciseness is marginal toward descriptive concepts (high word counts). 

\begin{table}[!tbp]
\caption{
Classification accuracy for HAM(skin disease images), BCCD(cytology images), and DR(fundus images) datasets using our generated concepts. LaBo (10K) runs the LaBo model until it reaches 10K epochs.
}
\label{tab:main_result}
\resizebox{1\columnwidth}{!}{
\begin{tabular}{c|c|c|c|c|c|c|c|c|c|c|c|c|c|c}
\toprule
&  \multicolumn{4}{c|}{Non-interpretable models} & \multicolumn{10}{c}{Interpretable CBM models} \\
\midrule
\multirow{2}{*}{Method} & Linear  & + Backbone  & +LoRA & Linear-probe & Label-free  & \multicolumn{3}{c|}{\multirow{2}{*}{LaBo \cite{yang2023language}}} & \multicolumn{3}{c|}{LaBo} & \multicolumn{3}{c}{AdaCBM }\\

& CLS & FT &  \cite{hu2022lora} & \& FT \cite{kumar2022fine} & CBM \cite{oikarinen2023label} & \multicolumn{3}{c|}{} & \multicolumn{3}{c|}{(10K) \cite{yang2023language}} & \multicolumn{3}{c}{(ours)}\\
\midrule
Concept  & \multirow{2}{*}{-} & \multirow{2}{*}{-} & \multirow{2}{*}{-} & \multirow{2}{*}{-} & \multirow{2}{*}{-} & \multicolumn{3}{c|}{\multirow{2}{*}{Submodular}} &\multicolumn{3}{c|}{\multirow{2}{*}{Submodular}} &  
\multicolumn{3}{c}{Concept Utility}\\
Selection &   &  &  &  &  & \multicolumn{3}{c|}{} &\multicolumn{3}{c|}{} &  
\multicolumn{3}{c}{Selection (ours)}\\
\midrule
$k$ & - & - & - & - & - & 10 & 20 & 50 & 10 & 20 & 50 & 10 & 20 & 50 \\
\midrule
HAM & 80.9$_{\pm 0.6}$ & 67.9$_{\pm 0.6}$ & 78.8$_{\pm 0.6}$ & 80.7$_{\pm 0.4}$ & 78.9$_{\pm 0.3}$ & 73.0$_{\pm 0.1}$ & 74.1$_{\pm 0.5}$ & 74.9$_{\pm 0.5}$ & 75.6$_{\pm 0.9}$ & 77.7$_{\pm 0.5}$ & 80.3$_{\pm 0.7}$ & 82.8$_{\pm 0.5}$ & \textbf{82.8$_{\pm 0.3}$} & 81.9$_{\pm 0.9}$\\
\midrule
BCCD & \bf 74.5$_{\pm 0.4}$ & 43.2$_{\pm 0.3}$ & 72.4$_{\pm 0.2}$ & 74.3$_{\pm 0.4}$ & 63.7$_{\pm 1.5}$ & 54.5$_{\pm 0.2}$ & 56.73$_{\pm 0.4}$ & 57.6$_{\pm 0.3}$ & 66.1$_{\pm 0.3}$ & 68.7$_{\pm 1.0}$ & 69.2$_{\pm 0.8}$ & 74.1$_{\pm 0.9}$ & 73.8$_{\pm 1.0}$ & 74.3$_{\pm 0.7}$\\
\midrule
DR & 78.0$_{\pm 0.1}$ & 73.7$_{\pm 0.2}$ & 71.1$_{\pm 0.4}$ & 77.5$_{\pm 0.2}$ & 75.7$_{\pm 0.1}$ & 74.6$_{\pm 0.1}$ & 75.4$_{\pm 0.1}$ & 75.8$_{\pm 0.1}$ & 74.7$_{\pm 0.1}$ & 76.0$_{\pm 0.1}$ & 77.2$_{\pm 0.1}$ & \textbf{78.3$_{\pm 0.1}$} & 78.3$_{\pm 0.1}$ & 78.2$_{\pm 0.1}$ \\
\bottomrule
\end{tabular}
}
\end{table}

\section{Experiments}
\noindent \textbf{Datasets.} We evaluate three publicly available medical image classification datasets encompassing different diseases and imaging modalities. %
\textbf{(1)} HAM10000 (HAM)~\cite{tschandl2018ham10000} is a skin disease dataset that comprises 10,015 dermatoscopic images collected over 20 years from two different sites, with 8,010 images available for training and 1,005 for testing. 
\textbf{(2)} BCCD~\cite{BCCD_Dataset} is a microscopic cytology image dataset to recognize four types of blood cells in a collection of 9,957 training images and 2,487 testing images. 
\textbf{(3)} DR~\cite{dugasdiabetic} is a fundus-based diabetic retinopathy (DR) dataset consisting of five grades from no to proliferative DR, comprising 35,126 and 53,576 images for training and testing respectively.

\noindent \textbf{Compared methods.} We compare our AdaCBM with the following methods: \textbf{(1)} LaBo \cite{yang2023language}, \textbf{(2)} Label-free CBM \cite{oikarinen2023label}, \textbf{(3)} fitting a linear classification (Linear CLS) model with a fixed backbone, \textbf{(4)} Linear CLS with backbone fine-tuning, \textbf{(5)} Linear-probe~\cite{kumar2022fine} followed by backbone fine-tuning, and \textbf{(6)} LoRA \cite{hu2022lora} fine-tuning of backbone with linear CLS.
We uniformly use CLIP's ViT-L/14 as the backbone for all compared methods except those explicitly mentioned.
Note that linear CLS is commonly known to produce better performance than CBMs~\cite{yuksekgonul2023posthoc,panousis2023sparse}, which is considered the `performance ceiling' in our comparison.

\noindent \textbf{Implementation details.} 
All experiments were conducted on a 48GB RTX A6000 GPU %
using PyTorch Lightning.
We employ an SGD optimizer with an initial learning rate of $5\times 10^{-4}$ with a linear decay to 1/100 of the initial value in a prefixed number of epochs. 
The weight decay was set to $1\times10^{-4}$.
We train 300 epochs for HAM and DR, and only 100 epochs for BCCD. 
All experiments were run five times and reported with the mean test accuracy from the last epoch.

\noindent \textbf{Doctor labeled concepts.}
A senior doctor manually reviewed the three medical tasks and provided 10 concepts from each of the color, shape, size, and texture aspects.
This is used as a comparison to our prompt-generated concepts. We found that the doctor-labeled concepts show close semantic similarity to the GPT-generated concepts (k=10) with the mean cosine similarity between them of 0.72, 0.66, and 0.70 for HAM, BCCD, and DTR respectively.

\begin{figure}[!hbp]
    \centering
    \resizebox{\columnwidth}{!}{
    \begin{tabular}{c}
         \includegraphics[width=\columnwidth]{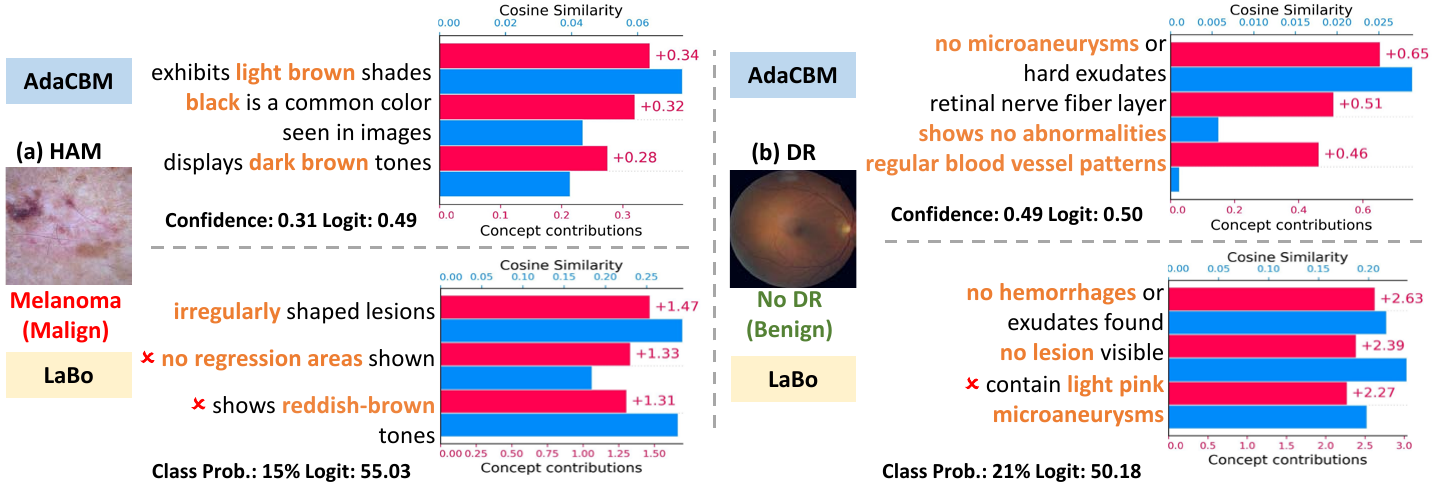}
    \end{tabular}
    }
    \caption{A comparison of AdaCBM and LaBo generated interpretation with the top-3 contributed concepts where both models predict the correct class. The {\color{red}red} and {\color{blue}blue} colored bars show the concept contribution $\mathbf{W}_{ji} s_j$ and the cosine similarity $\hat{\mathbf{x}} \cdot \hat{\mathbf{t}}$. Clinical-relevant phrases are highlighted in orange color. \ding{55} indicates a concept that is not initially selected for the predicted class. }
    \label{fig:da_cbm_example}
\end{figure}

\begin{figure*}[!tbp]
\centering
\includegraphics[width=1\textwidth]{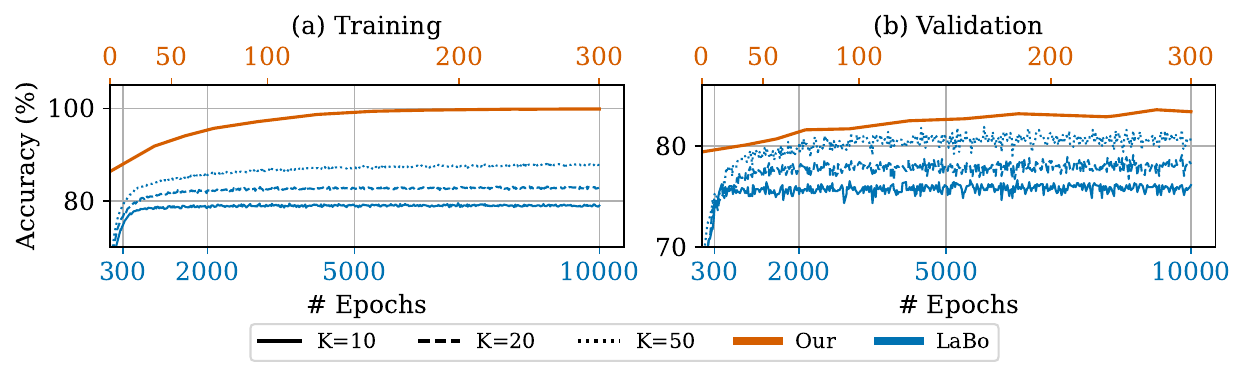}
\caption{Training and validation accuracy comparison of LaBo (10K) and our AdaCBM for $k \in \{10, 20, 50\}$ concepts (distinguished by line styles). 
Due to significant training length differences, we show AdaCBM in {\color{orange}orange} and LaBo (10k) in {\color{blue}blue} colored axis.
}
\label{fig:accuracy_curve}
\end{figure*}

\noindent \textbf{Result comparison.} We show the experiment results in Table~\ref{tab:main_result}.
The interpretation examples are shown in Fig.~\ref{fig:da_cbm_example}. Our observations are the following. 
\textbf{(1)} Regardless of the number of concepts $k = \frac{K}{n}$, our AdaCBM produces a comparable classification performance to Linear CLS. 
In contrast, all other methods show some degrees of performance deterioration, especially LaBo which has a larger deterioration rate for smaller $k$.
\textbf{(2)} The commonly used backbone fine-tuning and the recent LoRA finetuning strategies perform worse than linear CLS.
In comparison, our AdaCBM shows a better performance without fine-tuning the CLIP backbone. 
This also means AdaCBM can use pre-computed backbone features, significantly reducing the training complexity.
\textbf{(3)} Fig.~\ref{fig:da_cbm_example} demonstrates that our AdaCBM is constrained to use contributions from concepts selected for each class, but LaBo, although implements a sparse concept-weight contribution promotion design, accepting large and positive concept contribution from other classes.
\textbf{(4)} Fig.~\ref{fig:accuracy_curve} indicates that LaBo suffers from convergence issues given it optimizes only the post-CBM parameters $\mathbf{W}$ and $\mathbf{b}$.
In comparison, our AdaCBM efficiently meditates the domain gap given a comparatively larger modeling capacity is introduced to the model.

\noindent \textbf{Ablation studies.} 
\label{sec:ablation}
\noindent \textbf{Table~\ref{tab:ablation1}-(1) concept generation.} AdaCBM consistently performs better than LaBo using either our GPT-4 prompt-generated or doctor-labeled concepts, showing AdaCBM is robust to the concept generation method.
A qualitative comparison of the prompt-generated and doctor-labeled concepts is in Fig.~1 in the supplementary material.
\textbf{Table~\ref{tab:ablation1}-(2) Adapter module's position.} %
The ablation of the adapter module's placement (\ie inserted after the image encoder, text encoder, or both) confirms that changing text embeddings is not ideal because these image classification datasets do not contain text to meaningfully update the text embedding module, and do so will lead to lower classification performance.
\textbf{Table~\ref{tab:ablation1}-(3) Adapter module's structure.}
The comparison shows the peak performance is at 2-layer for HAM and 1-layer for BCCD and DR, which shows further addition of the adapter's capacity (\ie 4 layers and beyond) overfit to the downstream task.
\textbf{Supplementary Table 1-Right concept selection.} This ablation shows that AdaCBM is also robust to the use of other concept selection methods (\ie Submodular~\cite{yang2023language} from LaBo or Label-free CBM's~\cite{oikarinen2023label}). 
\textbf{Supplementary Table~2-(1) importance of geometrically represented quantities.}
We test the importance of each quantity by replacing them with 1 during the inference to inhibit their participation in the model decision. 
The norm inhibitions show marginal performance change, suggesting limited discriminative power.
In contrast, cosine similarity inhibition largely affects the performance. 
Hence we show both concept contribution and cosine similarity in Fig.~\ref{fig:da_cbm_example}.
\textbf{Supplementary Table 2-(2) LLM variants.} Our AdaCBM shows robust performance across both GPT-3/4 generated concepts.
\textbf{Supplementary Table 2-(3) Backbone variants.} This ablation shows that the representation power of a backbone ties to the CBM classification performance, thus backbone fine-tuning should be cautiously carried out to not overfit the downstream task.
\nocite{biomedclip,plip}

\begin{table}[t]
    \centering
    \caption{Ablation study on (1) concept generation using our GPT-4 prompt generated or doctor-labeled concepts, both with our concept utility selection method; (2) Adapter module's positioning; and (3) Adapter module's structure. All results are with $k=10$.}
    \resizebox{0.8\textwidth}{!}{
    \begin{tabular}{c||c|c|c|c||c|c|c||c|c|c}
    \toprule
    	&	\multicolumn{4}{c||}{(1) Concept Generation}							&	\multicolumn{3}{c||}{(2) Adapter  Module}					&	\multicolumn{3}{c}{(3) \# layers in}					\\ \cmidrule{2-5} 
Dataset	&	\multicolumn{2}{c|}{LaBo}			&	\multicolumn{2}{c||}{AdaCBM}			&	\multicolumn{3}{c||}{Placed After}					&	\multicolumn{3}{c}{Adapter  Module}					\\ \cmidrule{2-11} 
	&	Doctor	&	Prompt	&	Doctor	&	Prompt	&	Ima. Enc.	&	Text Enc.	&	Both	&	1	&	2	&	4	\\ \midrule
HAM	&	73.7	&	73.8	&	\bf 83.1	&	82.8	&	\bf 82.8	&	80.2	&	82.6	&	81.6	&	\bf 82.8	&	82.6	\\ \midrule
BCCD	&	54.7	&	59.6	&	\bf 74.3	&	74.1	&	\bf 74.1	&	70.8	&	71.7	&	\bf 74.1	&	72.6	&	73.1	\\ \midrule
DR	&	74.5	&	74.6	&	\bf 78.4	&	78.3	&	\bf 78.3	&	77.4	&	76.9	&	\bf 78.3	&	77.6	&	76.2	

    \\ \bottomrule
    \end{tabular}
    }
    \label{tab:ablation1}
\end{table}

\section{Conclusion}
Concept bottleneck models (CBMs) are interpretable deep learning classification models that face performance degradation challenges.
Here, we propose an adaptive CBM model to address the performance issue. The core contribution of our work is a fresh perspective on CBMs, examining their geometrical representation and treating CBMs as linear classification systems. This novel approach reveals an alternative way to increase model capacity, addressing the fine-tuning requirements for the medical domain, while preserving the interpretability that CBMs offer, an important issue that is currently overlooked by current CBM research. Future work will look into improving the clinical and semantic accuracy of the text concepts.

\begin{credits}
\subsubsection{\discintname}
The authors have no competing interests to declare that are relevant to the content of this article.
\end{credits}

\bibliographystyle{splncs04}
\bibliography{Paper-3895}

\begin{thebibliography}{10}
\providecommand{\url}[1]{\texttt{#1}}
\providecommand{\urlprefix}{URL }
\providecommand{\doi}[1]{https://doi.org/#1}

\bibitem{adebayo2018sanity}
Adebayo, J., Gilmer, J., Muelly, M., Goodfellow, I., Hardt, M., Kim, B.: Sanity checks for saliency maps. Advances in neural information processing systems  \textbf{31} (2018)

\bibitem{brown2020language}
Brown, T., Mann, B., Ryder, N., Subbiah, M., Kaplan, J.D., Dhariwal, P., Neelakantan, A., Shyam, P., Sastry, G., Askell, A., et~al.: Language models are few-shot learners. Advances in neural information processing systems  \textbf{33},  1877--1901 (2020)

\bibitem{brunese2020explainable}
Brunese, L., Mercaldo, F., Reginelli, A., Santone, A.: Explainable deep learning for pulmonary disease and coronavirus covid-19 detection from x-rays. Computer Methods and Programs in Biomedicine  \textbf{196},  105608 (2020)

\bibitem{dugasdiabetic}
Dugas, E., Jared, J., Cukierski, W.: Diabetic retinopathy detection (2015). URL https://kaggle.com/competitions/diabetic-retinopathy-detection

\bibitem{espinosa2022concept}
Espinosa~Zarlenga, M., Barbiero, P., Ciravegna, G., Marra, G., Giannini, F., Diligenti, M., Shams, Z., Precioso, F., Melacci, S., Weller, A., et~al.: Concept embedding models: Beyond the accuracy-explainability trade-off. NIPS  (2022)

\bibitem{havasi2022addressing}
Havasi, M., Parbhoo, S., Doshi-Velez, F.: Addressing leakage in concept bottleneck models. NIPS  (2022)

\bibitem{hu2022lora}
Hu, E.J., Shen, Y., Wallis, P., Allen-Zhu, Z., Li, Y., Wang, S., Wang, L., Chen, W.: Lo{RA}: Low-rank adaptation of large language models. In: ICLR (2022)

\bibitem{plip}
Huang, Z., Bianchi, F., Yuksekgonul, M., Montine, T.J., Zou, J.: A visual--language foundation model for pathology image analysis using medical twitter. Nature Medicine pp. 1--10 (2023)

\bibitem{koh2020concept}
Koh, P.W., Nguyen, T., Tang, Y.S., Mussmann, S., Pierson, E., Kim, B., Liang, P.: Concept bottleneck models. In: International conference on machine learning. pp. 5338--5348. PMLR (2020)

\bibitem{kumar2022fine}
Kumar, A., Raghunathan, A., Jones, R., Ma, T., Liang, P.: Fine-tuning can distort pretrained features and underperform out-of-distribution. ICLR  (2022)

\bibitem{liao2020multi}
Liao, L., Zhang, X., Zhao, F., Lou, J., Wang, L., Xu, X., Zhang, H., Li, G.: Multi-branch deformable convolutional neural network with label distribution learning for fetal brain age prediction. In: 2020 IEEE 17th International Symposium on Biomedical Imaging (ISBI). pp. 424--427. IEEE (2020)

\bibitem{lipton2017doctor}
Lipton, Z.C.: The doctor just won't accept that! arXiv preprint arXiv:1711.08037  (2017)

\bibitem{maas2013rectifier}
Maas, A.L., Hannun, A.Y., Ng, A.Y., et~al.: Rectifier nonlinearities improve neural network acoustic models. In: Proc. icml. vol.~30, p.~3. Atlanta, GA (2013)

\bibitem{oikarinen2023label}
Oikarinen, T., Das, S., Nguyen, L.M., Weng, T.W.: Label-free concept bottleneck models. In: International Conference on Learning Representations (2023)

\bibitem{panousis2023sparse}
Panousis, K.P., Ienco, D., Marcos, D.: Sparse linear concept discovery models. In: Proceedings of the IEEE/CVF International Conference on Computer Vision. pp. 2767--2771 (2023)

\bibitem{radford2021learning}
Radford, A., Kim, J.W., Hallacy, C., Ramesh, A., Goh, G., Agarwal, S., Sastry, G., Askell, A., Mishkin, P., Clark, J., et~al.: Learning transferable visual models from natural language supervision. In: International conference on machine learning. pp. 8748--8763. PMLR (2021)

\bibitem{rudin2019stop}
Rudin, C.: Stop explaining black box machine learning models for high stakes decisions and use interpretable models instead. Nature machine intelligence  \textbf{1}(5),  206--215 (2019)

\bibitem{selvaraju2017grad}
Selvaraju, R.R., Cogswell, M., Das, A., Vedantam, R., Parikh, D., Batra, D.: Grad-cam: Visual explanations from deep networks via gradient-based localization. In: IEEE Intl. Conf. on Computer Vision. pp. 618--626 (2017)

\bibitem{BCCD_Dataset}
Shenggan: Bccd dataset. URL https://github.com/Shenggan/BCCD\_Datase  (2017)

\bibitem{srivastava2014dropout}
Srivastava, N., Hinton, G., Krizhevsky, A., Sutskever, I., Salakhutdinov, R.: Dropout: a simple way to prevent neural networks from overfitting. The journal of machine learning research  \textbf{15}(1),  1929--1958 (2014)

\bibitem{tschandl2018ham10000}
Tschandl, P., Rosendahl, C., Kittler, H.: The ham10000 dataset, a large collection of multi-source dermatoscopic images of common pigmented skin lesions. Scientific data  \textbf{5}(1), ~1--9 (2018)

\bibitem{xie2020mutual}
Xie, Y., Zhang, J., Xia, Y., Shen, C.: A mutual bootstrapping model for automated skin lesion segmentation and classification. IEEE transactions on medical imaging  \textbf{39}(7),  2482--2493 (2020)

\bibitem{yang2023language}
Yang, Y., Panagopoulou, A., Zhou, S., Jin, D., Callison-Burch, C., Yatskar, M.: Language in a bottle: Language model guided concept bottlenecks for interpretable image classification. In: Proceedings of the IEEE/CVF Conference on Computer Vision and Pattern Recognition. pp. 19187--19197 (2023)

\bibitem{yuksekgonul2023posthoc}
Yuksekgonul, M., Wang, M., Zou, J.: Post-hoc concept bottleneck models. In: The Eleventh International Conference on Learning Representations (2023)

\bibitem{biomedclip}
Zhang, S., Xu, Y., Usuyama, N., Bagga, J., Tinn, R., Preston, S., Rao, R., Wei, M., Valluri, N., Wong, C., Lungren, M., Naumann, T., Poon, H.: Large-scale domain-specific pretraining for biomedical vision-language processing (2023)

\bibitem{zhou2016learning}
Zhou, B., Khosla, A., Lapedriza, A., Oliva, A., Torralba, A.: Learning deep features for discriminative localization. In: Proceedings of the IEEE conference on computer vision and pattern recognition. pp. 2921--2929 (2016)

\end{thebibliography}
\end{document}


\begin{algorithm}[!ht]
\DontPrintSemicolon
  \KwInput{$\mathcal{X}$ - list of image embeddings\\
  \quad\quad\quad$\mathcal{T}$ - list of text embeddings\\
  \quad\quad\quad$k$ - number of concepts to be selected for each class, $k = \frac{K}{n}$\\ 
  \quad\quad\quad$c$ - target class\\
  \quad\quad\quad$\gamma \in [0, 1]$ - threshold for Pearson's $r$}
  \KwOutput{$\mathcal{O}$ - selected text embeddings}
  $\mathcal{O} \leftarrow \{\}$ \tcp*{empty set}
   \While{$|\mathcal{O}| < k$}
   {
   	$\mathbf{t} \leftarrow \text{argmax}_\mathbf{t \in \mathcal{T}} U(\mathbf{t}, c)$\;
        $\mathcal{O} \leftarrow \mathcal{O} \cup \{\mathbf{t}\}$\;
        $\mathcal{T} \leftarrow \mathcal{T} \setminus \{\mathbf{t}\}$\;
        $\mathcal{R} \leftarrow \{\mathbf{t}' | \text{ abs}\big(\rho(\mathbf{t}, \mathbf{t}')\big) > \gamma, \mathbf{t}' \in \mathcal{T}\}$ \; \tcp*{$\rho(., .)$ denotes Pearson's $r$}
        \If{$k - |\mathcal{O}| <= |\mathcal{T} \setminus \mathcal{R}|$}
        {
        $\mathcal{T} \leftarrow T \setminus \mathcal{R}$\;
        }\Else
        {
        $\mathcal{A} \leftarrow \text{utility\_selection}(\mathcal{X}, \mathcal{T}, k - |\mathcal{O}|, c, \gamma + 0.1)$
        $\mathcal{O} \leftarrow \mathcal{O} \cup \mathcal{A}$\;
        }
    }
    return $\mathcal{O}$
\caption{utility\_selection($\mathcal{X}$, $\mathcal{T}$, $k$, $c$, $\gamma$)}
\label{alg:concept_selection}
\end{algorithm}

\begin{table}[!htbp]
\caption{(Left) An example of concept selection outcome by using Algorithm \ref{alg:concept_selection} on the HAM dataset with $k=50$ selected concepts from a total of GPT-4 generated 760 concepts. (Right) Comparison of concept selection method for HAM dataset using our concept
generation method. }
\scalebox{0.67}{
\begin{minipage}{.6\linewidth}
\begin{tabular}{c|c|c|c|c|c}
\toprule
Avg. word len.  & Color & Shape & Size & Texture & Total \\
\midrule
4.4 & 22 & 36 & 21 & 36 & 115\\
6.0 & 23 & 26 & 18 & 38 & 105\\
8.7 & 30 & 33 & 28 & 39 & 130\\
\midrule
Total & 75 & 95 & 67 & 113 & 350\\
\bottomrule
\end{tabular}
\end{minipage}
\hfill
\begin{minipage}{.5\linewidth}
\begin{tabular}{c|c|c|c|c|c|c|c}
\toprule
Concept Selection $\rightarrow$ & \multicolumn{3}{c|}{Concept Utility (ours)} & \multicolumn{3}{c|}{Submodular [23]} &  Label-free \\
\cline{1-7}
CBM $\downarrow$ \textbackslash ~$k\rightarrow$ & $10$ & $20$ & $50$ & $10$ & $20$ & $50$ & CBM [14]\\
\hline
LaBo & 73.8 & 75.3 & 76.8 & 73.0 & 73.6 & 74.7 & 72.6\\
\midrule
AdaCBM & 82.8 & 82.8 & 81.9 & 82.9 & 82.9 & 82.1 & 81.8\\
\bottomrule
\end{tabular}
\end{minipage}
}
\label{tab:supp1}
\end{table}

\begin{figure}[!htbp]
\centering
\caption{Top-5 semantically similar concept pairs for a ``Dermatofibroma" case. We show the cosine similarity between each pair using the CLIP text encoder-produced embeddings. Even if semantically similar, each pair's score is as high as we expect. However, our AdaCBM is robust to concept generation, which can achieve high performance on both concept types as shown in Table 2-(1) in the main manuscript.}
\scalebox{1}{
\resizebox{1\columnwidth}{!}{
\begin{tabular}{c|c|c|c}
\toprule
Example Image & GPT Generated Concepts & Cos. Sim. & Doctor-labeled Concepts\\
\hline
\multirow{6}{*}{\includegraphics[width=0.15\columnwidth]{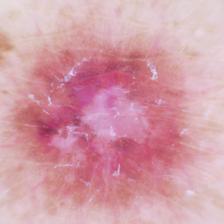}}
& 1. rarely, it might show up as a dome-shaped 
& \multirow{2}{*}{0.81}
& 1. usually dome-shaped\\
& bump on the skin 
& 
& \\
\cline{2-4}
& 2. may appear as oval shape 
& 0.71
& 2. generally round but can be oval\\
\cline{2-4}
& 3. appear as a light pink hue 
& 0.69
& 3. occasionally can be pink or red\\
\cline{2-4}
& 4. may exhibit central hardening 
& 0.55
& 4. can appear indurated or hardened\\
\cline{2-4}
& 5. may reach up to 10mm in diameter
& 0.49
& 5. size usually ranges from 3 to 10 mm\\
\bottomrule
\end{tabular}
}
}
\label{tab:doctor}
\end{figure}

\begin{table}[t]
    \centering
    \caption{Ablation study of the proposed AdaCBM model on (1) the importance of the geometrically represented quantities in terms of contribution to accuracy; (2) GPT-3/-4 generated concepts; (3) AdaCBM trained with different backbones. All results are generated on the HAM dataset. The Baseline, GPT-4, and ViT-L/14 columns are identical as they are named to the different aspects of the same baseline AdaCBM model in Table~1 in the main manuscript.}
    \resizebox{\textwidth}{!}{
    \begin{tabular}{c||c||c|c|c|c||c|c||c|c|c|c|c}
    \toprule
    	&	\multicolumn{5}{c||}{(1) Importance of the Geometrically}									&	\multicolumn{2}{c||}{\multirow{2}{*}{(2) LLM}}			&	\multicolumn{5}{c}{(3) Backbones}									\\ \cmidrule{9-13} 
$k$	&	\multicolumn{5}{c||}{Represented Quantities}									&	\multicolumn{2}{c||}{}			&	\multicolumn{3}{c|}{CLIP}					&	BioMedCLIP 	&	PLIP 	\\ \cmidrule{2-11} 
	&	Baseline	&	$\lVert\mathbf{x}\rVert =1$	&	$\lVert\mathbf{t}\rVert =1$	&	$\lVert\mathbf{x}\rVert\lVert\mathbf{t}\rVert = 1$	&	$\hat{\mathbf{x}}\cdot\hat{\mathbf{t}} = 1$ 	&	GPT-3	&	GPT-4	&	ViT-L/14	&	ViT-B/32	&	ResNet-50	&	[25]	&	[8]\\ \midrule
10	&	82.8	&	\bf 82.9	&	82.8	&	82.8	&	66.8	&	\bf 82.9	&	82.8	&	\bf 82.8	&	79.1	&	77.4	&	67.8	&	82.5	\\ \midrule
20	&	82.8	&	82.8	&	\bf 83.2	&	82.6	&	3.3	&	\bf 82.8	&	\bf 82.8	&	82.8	&	79.2	&	78.8	&	70.7	&	\bf 82.9	\\ \midrule
50	&	81.9	&	\bf 82.6	&	78.8	&	78.1	&	1.2	&	\bf 82.4	&	81.9	&	\bf 81.9	&	79.3	&	81.3	&	71.6	&	81.8	

    \\ \bottomrule
    \end{tabular}
    }
    \label{tab:ablation2}
\end{table}